%% file: mes_whi.tex
\documentclass{article}
\usepackage{times}
\usepackage{graphicx}
\usepackage{amssymb}
\usepackage{amsmath}
\usepackage{subfigure}
\usepackage{watermark}

\usepackage{natbib}

\usepackage{algorithm}
\usepackage{algorithmic}

\usepackage{hyperref}

\usepackage[accepted]{whi2016}

\include{mathmacros}

\icmltitlerunning{A Model Explanation System: Latest Updates and Extensions}

\newcommand{\expl}{E}
\newcommand{\explr}{g}
\newcommand{\nexplr}{M}
\newcommand{\allE}{\set{E}}

\newcommand{\Estar}{{E^*}}
\newcommand{\decfunc}{{f}}

\newcommand{\mcsample}{v}

\newcommand{\U}{\bigcup}

\newcommand{\logicaland}{\wedge}

\newcommand{\logicalnot}{\neg}

\newcommand{\step}{u}
\newcommand{\ZOL}{\ell}
\newcommand{\surrloss}{\phi}

\newcommand{\argeq}{=}

\begin{document}

\twocolumn[
\icmltitle{A Model Explanation System: Latest Updates and Extensions}

\icmlauthor{Ryan Turner}{ryan.turner@ngc.com}
\icmladdress{Northrop Grumman Corporation}

\icmlkeywords{model interpretability, model criticism, visualization, face recognition, credit scoring, fraud detection, anomaly detection, SVM, sparsity, decision tree}

\vskip 0.3in
]


\begin{abstract}
We propose a general model explanation system (MES) for ``explaining'' the output of black box classifiers.
This paper describes extensions to~\citet{Turner2015}, which is referred to frequently in the text.
We use the motivating example of a classifier trained to detect fraud in a credit card transaction history.
The key aspect is that we provide explanations applicable to a \emph{single prediction}, rather than provide an interpretable set of parameters.
We focus on explaining positive predictions (alerts)\@.
However, the presented methodology is symmetrically applicable to negative predictions.
\end{abstract}

In many classification applications, but especially in fraud detection, there is an expectation of false positives.
Alerts are given to a human \emph{analyst} before any further action is taken.
Such problems are sometimes referred to as ``anomaly detection.''
Analysts often insist on understanding ``why'' there was an alert, since an opaque alert makes it difficult for them to proceed.
Analogous scenarios occur in computer vision, credit risk, spam detection, etc.

Furthermore, the MES framework is useful for model criticism.
In the world of generative models, practitioners often generate synthetic data from a trained model to get an idea of ``what the model is doing''~\citep{Gelman1996}.
Our MES framework augments such tools.
As an added benefit, MES is applicable to completely nonprobabilistic black boxes that only provide hard labels.

\begin{figure}[t]
  \includegraphics[scale=1.2]{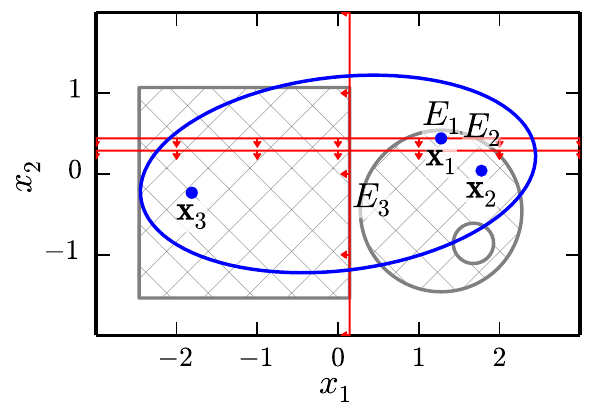}
  \vspace{-2mm}
  \caption{{\small
  Illustration of MES on toy classifier with test inputs $\vec x_1$, $\vec x_2$, and $\vec x_3$ (blue dots)\@.
  The classifier $f$ outputs 1 in the hatched regions and 0 elsewhere.
  The input density on the data is Gaussian (blue ellipse).
  The red boundaries are the respective explanations ($E_1$, $E_2$, and $E_3$) for each of the test inputs $\vec x$.
  The explanation $E_1$ for $\vec x_1$ is: $[\vec x_1]_2 \leq 0.5$.
  Note the red arrows that depict the $\leq$ relation.
  We also have $E_2$: $[\vec x_2]_2 \leq 0.25$; and $E_3$: $[\vec x_3]_1 \leq 0.15$.
  As most of the data comes from inside the blue ellipse, MES does not care that the explanations disagree with the classifier at the plot's extremities.
  Although this example is 2D, MES is applicable in high dimensions.}}
  \label{fig:funky example}
\end{figure}

\vspace{-1mm}
\paragraph{Example}
In the context of credit card fraud we may have feature vectors $\vec x$ containing the number of online transactions, the geographic distance traveled for in-person transactions, the number of novel merchants, and so on.
A simple example explanation is:
``Today, there were two in-person transactions in the USA, followed by \$1700 in country X.''
MES would output ``$(x_i \geq 2) \logicaland (x_j \geq 1700)$'' for the appropriate features $i$ and $j$.
We graphically depict MES on a separate illustrative example in Fig.~\ref{fig:funky example}.

\vspace{-1mm}
\paragraph{Explanation vs.~interpretability}
We adopt the paradigm where prediction accuracy is of paramount importance, but explanation is also important.
Therefore, we are not willing to give up any predictive accuracy for explanation.
Both machine learning and statistics have a long history of building models that are ``interpretable''; such as, (small) decision trees~\citep{Quinlan1986} and sparse linear models~\citep{Tibshirani1996}.
MES augments black boxes with explanations, as the best predictor may not be ``interpretable.''

Historically, this dilemma has created two distinct approaches:
1)~the ``interpretable'' models approach, common in scientific discovery/bioinformatics, and
2)~the accuracy-focused approach, common in computer vision with methods including deep learning, $k$-NNs, and support vector machines (SVMs)\@.
The downside of the interpretable approach is seen in machine learning competitions, where the winning methods are typically nonparametric, or have a very large number of parameters (e.g., deep networks)\@.

MES has elements of both approaches.
We do not aim to summarize how the model ``works in general''~\citet{Andrews1995}, but only seek explanations of individual cases.
Although the distinction is subtle, explanation is a much easier task than explaining an entire model.
MES is the first method to utilize this weaker requirement to augment black boxes with explanations without affecting accuracy.

\section{Formal setup}
\label{sec:Formal setup}

Consider a black box binary classifier $\decfunc$ that takes a feature vector $\vec x \in \set{X}=\R^D$ and provides a binary label: $\decfunc \in \set{X} \rightarrow \{0,1\}$.
In the introductory examples, explanations are Boolean statements about the feature vector.
In effect, an explanation $\expl$ is a function from $\set{X}$ to $\{0,1\}$.
The mapping $\Estar \in \set{X} \rightarrow \allE$ finds the best explanation from the set of possible explanations $\allE \subset \set{X} \rightarrow \{0,1\}$.
We also define that $\allE$ contains a ``null explanation'' $\expl_0(\vec x) := 1$.
Note that an explanation is either sufficiently simple to be in $\allE$ or not.
There is no other metric of ``explanation simplicity.''

\citet{Turner2015} formalized axioms on what properties a sensible explanation system $\Estar$ should have.
One possibility, that also has favorable computational properties, is an optimization over the following explanation \emph{quality score} $S$:
\begin{align}
  \Estar(\vec x) &\argeq \textstyle{\argmax_{\expl \in \allE}} S(\expl) \quad \textrm{s.t.} \quad \expl(\vec x) = 1\,, \label{eq:MI opt} \\
  S(\expl) &= P(\expl(\vec x') \given \decfunc(\vec x')=1) - P(\expl(\vec x') \given \decfunc(\vec x')=0)\,, \nonumber 
\end{align}
where $\decfunc$ and $\expl$ are deterministic functions; we are \emph{marginalizing} over the input distribution $p(\vec x')$.
Notably, $S$ is equivalent to the covariance: $S(\expl) \propto \cov{\expl,\decfunc}$.
Under this definition the null $\expl_0$ has score $S(\expl_0)=0$, and the true classifier $\decfunc$ has score $S(\decfunc)=1$.
Therefore, if the decision rule $\decfunc$ is in $\allE$, then it is preferable to any other explanation;
and the selected explanation $\expl = \Estar(\vec x)$ has a normalized quality score: $S(\expl) \in {[0,1]}$.
Also note that by construction, any explanation $\expl \neq \expl_0$ selected for explaining $\decfunc(\vec x)=1$ would be not be selected for the converse problem of trying to explain $\decfunc(\vec x)=0$.

\section{Score estimation with black box models}
\label{sec:Black box models}

This section reviews using simple Monte Carlo to approximate the optimization in~\eqref{eq:MI opt} with black box models.
We merely require the classifier $\decfunc$ be queryable at an arbitrary input $\vec x$ and that we can obtain samples from the input density $p(\vec x)$.
We allow for general explanation functions of the form $\explr_i \in \set{X} \rightarrow \R$:
\begin{align}
  \allE &= \textstyle{\U_{i=1}^\nexplr} \left\{\I\{\explr_i(\vec x) \leq a\},\, \forall a \in \R\right\}\,. \label{eq:general expl}
\end{align}
Explanations of the form \smash{$\I\{\explr_i(\vec x) \geq a\}$} are obtainable by including \smash{${\explr'}_{\!i}=-\explr_i$ in $\allE$}.
The \emph{axis aligned} explanations from Fig.~\ref{fig:funky example} are recovered using $\explr(\vec x) = \pm x_i$, yielding
\smash{$\allE \!=\! \U_{i=1}^D \!\{\I\{x_i \lesseqgtr a\}, \forall a \in \R\}$}.
Alternatively, we may have a predefined set of linear decision functions that are reasonable explanations:
\smash{$\explr_i(\vec x) = \vec w_i\T \vec x + b_i$}.

The optimization to find the best explanation is done as follows:
For each explanation function $\explr_i$, we utilize the output of a precomputation phase to efficiently find the optimal threshold $\hat{a}$ and its corresponding score.
We then compare the optimized scores for each explanation function $\explr_i$ and report the function $\explr_i$ (and corresponding threshold $\hat{a}$) with the highest score.
\citet{Turner2015} showed that using Algo.~\ref{alg:MES MC Precomputation} for precomputation requires
$n = \left\lceil 8 \log(4 \nexplr/\delta) / \epsilon^2 \right\rceil$
MC samples to obtain score suboptimality $\epsilon$ with confidence $\delta$.

The precomputation phase, Algo.~\ref{alg:MES MC Precomputation}, is based on finding the \emph{cumulative maximum} w.r.t.~$a$ of the estimated score function $\hat{S}$.
The $\max$ in Algo.~\ref{alg:MES MC Precomputation} is a tiebreaker so that $\hat{a}$ equals the largest $a$ of the set returned by the $\argmax$.
The computations to find \smash{$A_{1:\nexplr}$} are informally thought of as the best optimum so far scanning from $+\infty$ backwards.
After precomputation, we efficiently find the explanation for a test point $\vec x$ using Algo.~\ref{alg:Run MES}.

\begin{algorithm}[tb]
  \caption{MES MC Precomputation}
  \label{alg:MES MC Precomputation}
  \begin{algorithmic}
    \INPUT classifier $\decfunc$, input density $p$, $\explr_{1:M}$, accuracy~($\epsilon$, $\delta$)
    \STATE Find $n$ from $\epsilon$ and $\delta$
    \STATE Sample iid
    $\mcsample_{1:n}^0 \sample p(\vec x \given \decfunc=0)$ and
    $\mcsample_{1:n}^1 \sample p(\vec x \given \decfunc=1)$
    \FOR{$i=1$ {\bfseries to} $\nexplr$}
    \STATE $H_n, \,\, F_n \assign \textrm{ECDF}(\explr_i(\mcsample_{1:n}^0)), \,\, \textrm{ECDF}(\explr_i(\mcsample_{1:n}^1))$
    \STATE $\hat{S}_i \assign F_n - H_n$
    \STATE $A_i(z) \assign \max \argmax_{a \in [z, \infty)} \hat{S}_i(a)\,, \quad \forall z \in \R$
    \ENDFOR
    \OUTPUT step-based functions
    $\hat{S}_{1:\nexplr}$ and $A_{1:\nexplr}$
\end{algorithmic}
\end{algorithm}

\begin{algorithm}[tb]
  \caption{Run MES}
  \label{alg:Run MES}
  \begin{algorithmic}
    \INPUT test input $\vec x$, $\hat{S}_{1:\nexplr}$, and $A_{1:\nexplr}$
    \FOR{$i=1$ {\bfseries to} $\nexplr$}
    \STATE Saving threshold $a$ with best score so far:
    \STATE Try $a \assign A_i(\explr_i(x))$ and its score $\hat{S}_i(a)$
    \ENDFOR
    \OUTPUT best threshold $a$, index $i$, and score
  \end{algorithmic}
\end{algorithm}

\section{Extending to larger explanation spaces}
\label{sec:extended MES}

In Section~\ref{sec:Black box models} we reviewed the machinery for jointly choosing among $\nexplr$ explanation functions $\explr_{1:\nexplr}$ and a scalar threshold parameter $a \in \R$.
In this section we propose \emph{extended MES}, which maximizes the score $S$ with respect to some continuous free parameters $\vec\theta$ of the explanation $\explr$.
For instance, Section~\ref{sec:Black box models} mentions using linear decision functions as explanations.
In this section we assume explanations of the general form:
\begin{align}
  \allE &= \left\{\I\!\left\{\explr(\vec x;\vec\theta) \leq a\right\},\, \forall a \in \R,\, \forall \vec\theta\right\}\,, \label{eq:free param expl}
\end{align}
where $\explr$ is now parameterized by $\vec\theta$ rather than a discrete index $i$.
In the case of linear explanations $\vec\theta = \vec w \in \R^D$.
We now have to optimize the score~\eqref{eq:MI opt} with respect to a free vector parameter $\vec\theta$.
To do this efficiently we put the objective in the form of an expected loss.
This enables us to employ learning theoretic results that replace the optimization with a convex surrogate.

First, we find it convenient to rewrite the explanations as:
\begin{align}
  \I\!\left\{\explr(\vec x;\vec\theta) \leq a\right\}
  &= \step(a - \explr(\vec x;\vec\theta)) = \step(\tilde{\explr}(\vec x;\tilde{\vec\theta}))\,, \label{eq:define augmented expl} \\
  \tilde{\explr}(\vec x;\tilde{\vec\theta}) &:= a - \explr(\vec x;\vec\theta)\,, \quad
    \tilde{\vec\theta}\TT :=
    \begin{bmatrix}
      \smash{\vec\theta\T} & a
    \end{bmatrix}\,,
    \nonumber
\end{align}
where $\step(\cdot)$ is the unit step function.
Since the explanation space $\allE$ is now parameterized by \smash{$\tilde{\vec\theta}$},~\eqref{eq:MI opt} is equivalent to:
\begin{align}
  \theta^*(\vec x) &\argeq \textstyle{\argmin_{\tilde{\vec\theta}}} \E_{\vec x'}[\step(\tilde{\explr}(\vec x';\tilde{\vec\theta})) \given \logicalnot \decfunc]
    \!-\! \E_{\vec x'}[\step(\tilde{\explr}(\vec x';\tilde{\vec\theta})) \given \decfunc] \nonumber \\
  &\textrm{s.t.} \quad \tilde{\explr}(\vec x;\tilde{\vec\theta}) \geq 0\,, \label{eq:param in theta}
\end{align}
where $\theta^*(\vec x)$ are the best parameters \smash{$\tilde{\vec\theta}$} for explaining $\vec x$.
By defining a ``class rebalanced'' version of $p$, we achieve the expected loss formulation:
\begin{align}
  &\theta^*\!(\vec x) \!\argeq\! \smash{\textstyle{\argmin_{\tilde{\vec\theta}}}}\, \E_{p'}[\ZOL(y\, \tilde{\explr}(\vec x';\tilde{\vec\theta}))]
    \,\, \textrm{s.t.} \,\, \tilde{\explr}(\vec x;\tilde{\vec\theta}) \geq 0\,, \nonumber \\
  &p'(\vec x',y) := p(\vec x' \given 2\decfunc(\vec x') - 1 = y) \, \tfrac{1}{2} \I\{y \in \{{-1},1\}\}\,, \nonumber
\end{align}
where we have manipulated different forms of the zero-one loss $\ZOL(x) := \step(-x)$:
$|\step(\hat{\decfunc}) - \decfunc| = \ZOL((2\decfunc - 1) \hat{\decfunc}) = \ZOL(y \hat{\decfunc})$ for some prediction \smash{$\hat{\decfunc} \in \R$}.
Although this objective can be estimated with MC samples from $p'$, the resulting function is multivariate and discontinuous.
This makes direct optimization problematic.
However, \citet{Bartlett2006} showed zero-one loss objectives can be solved by replacing $\ZOL$ with a convex surrogate loss $\surrloss \in \R \rightarrow \R^+$ such as the \emph{hinge loss} or log-logistic:
\begin{align}
  \theta^*\!(\vec x) \!\argeq\! \smash{\textstyle{\argmin_{\tilde{\vec\theta}}}}\, \E_{p'}[\surrloss(y\, \tilde{\explr}(\vec x';\tilde{\vec\theta}))]
    \,\, \textrm{s.t.} \,\, \tilde{\explr}(\vec x;\tilde{\vec\theta}) \geq 0\,. \nonumber
\end{align}
If we take a large number of MC samples, the resulting parameter estimates have asymptotically minimal risk.

Although it is possible to solve for $\theta^*(\vec x)$ directly by constrained optimizing, we take the ``poor man's'' approach of putting the constraint (\smash{$\tilde{\explr}(\vec x;\tilde{\vec\theta}) \geq 0$}) in the objective.
This has the practical advantage of allowing us to use existing (highly optimized) software modules.
We modify our objective as follows using $\gamma \in (0,0.5)$:
\begin{align}
  &\theta^*(\vec x) \argeq \textstyle{\argmin_{\tilde{\vec\theta}}} \gamma \E_{p'}[\ZOL(y\, \tilde{\explr}(\vec x';\tilde{\vec\theta}))]
    + (1\!-\!\gamma) \ZOL(\tilde{\explr}(\vec x;\tilde{\vec\theta})) \nonumber \\
  &= \textstyle{\argmin_{\tilde{\vec\theta}}} \E_{p''}[\ZOL(y\, \tilde{\explr}(\vec x';\tilde{\vec\theta}))]\,, \\
  &p''(\vec x',y) := (1\!-\!\gamma) \I\{y=1\} \delta_{\vec x}(\vec x') + \gamma p'(\vec x',y)\,, \label{eq:ppp def}
\end{align}
where \smash{$\delta_{\vec x}(\cdot)$} is a Dirac delta centered at $\vec x$.
In the case of linear explanations we have
\smash{$\tilde{\explr}(\vec x;\tilde{\vec\theta}) = \tilde{\vec\theta}\TT \tilde{\vec x}$}, where we have defined
\smash{$\tilde{\vec x}\TT :=
\begin{bmatrix}
  \vec x\TT & 1
\end{bmatrix}$}.
This gives us a final objective of:
\begin{align}
  \theta^*(\vec x) &\argeq \textstyle{\argmin_{\tilde{\vec\theta}}} \sum_{i=1}^n \surrloss(y_i\, \tilde{\vec\theta}\TT \tilde{\vec x}_i)\,,
    \quad (\vec x_i, y_i) \sample p''\,.
    \nonumber
\end{align}
When $\surrloss$ is the log-logistic we find \smash{$\tilde{\vec\theta}$} by applying logistic regression to MC samples $\data := (\vec x_{1:n}, y_{1:n})$.
Likewise, when $\surrloss$ is the hinge loss we use a linear SVM\@.
Finally, we map $\tilde{\vec\theta}$ back to $(\vec w, b)$ for a linear explanation using~\eqref{eq:define augmented expl}.

\begin{figure*}
  \centering
  \includegraphics[width=0.95\hsize]{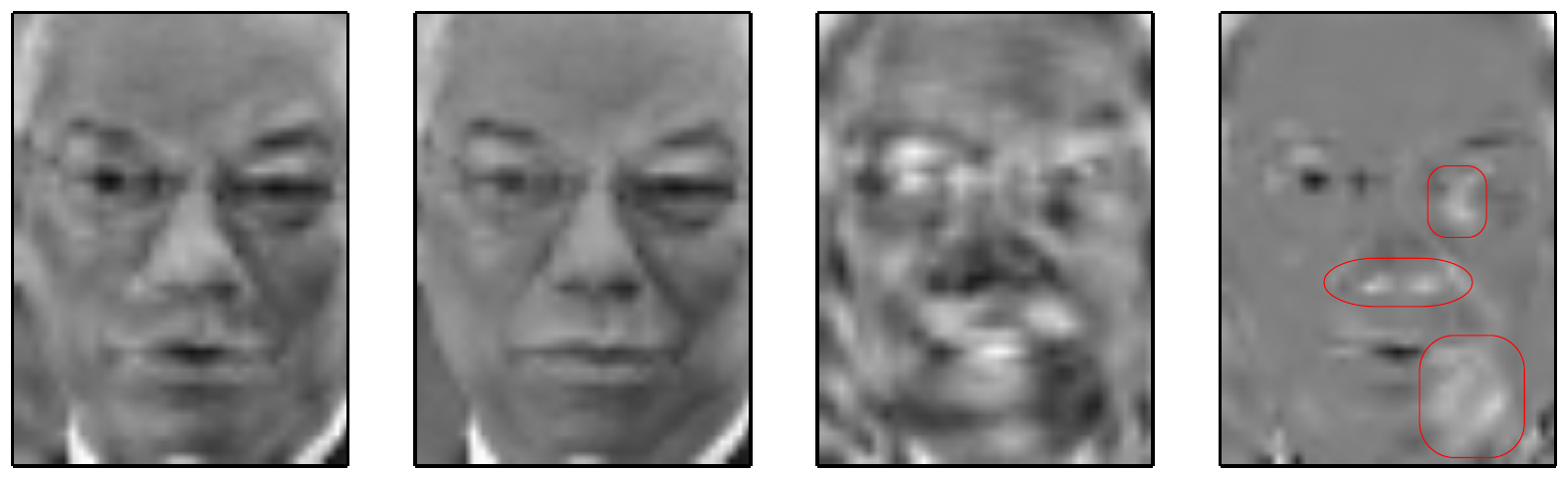}
  \caption{{\small
  Example of MES explaining a correct prediction of Powell by the (nonlinear) SVM classifier.
  This example used extended MES (Algo.~\ref{alg:Extended MES} followed by Algos.~\ref{alg:MES MC Precomputation} and~\ref{alg:Run MES}) to learn the optimal linear explanation.
  We subtract out the explanation face (\textbf{right}) from the (mean removed) original (\textbf{left}) to make the image on the \textbf{far left}.
  In these images: $\textrm{gray}=0$, $\textrm{white}>0$, and $\textrm{black}<0$\@.
  The product image (\textbf{far right}) is the Hadamard product of the original face and the explanation face.
  Here, the explanation is that the product image has net white balance $1.6\% > 0.5\%$, with a score of $S=0.865$.
  We have added the red annotations as cues to the reader on the important areas.
  \textbf{Technical~details:}
  The above images are created as follows:
  Let $\vec x$ be the mean removed input face (left) reshaped as a vector.
  This is transformed by PCA to get $\vec x_{\textrm{PCA}} := \mat C \vec x$, where $\mat C$ is the principal component matrix.
  The explanation is: \smash{$\vec w\T \vec x_{\textrm{PCA}} > a$}.
  Thus we set the right image to be \smash{$\vec x_E := \mat C\T \vec w$}.
  We then set the far right image to be $\vec x_H := \mat x_E \hada \vec x$.
  Then the explanation becomes: \smash{$\vec x_E \cdot \vec x = \sum \vec x_H > a$}.
  We set the corrected image to be \smash{$\vec x_F := \vec x - \alpha \vec x_E / ||\vec x_E||^2$}.
  When applying the explanation to the corrected image we get: $\vec x_E \cdot \vec x_F = \vec x_E \cdot \vec x - \alpha$.
  Thus, by setting \smash{$\alpha > \sum \vec x_H - a$}, the explanation is false: $\expl(\vec x_F)=0$.
  Here, $\alpha=2$.
  }}
  \label{fig:linear face opt}
  \vspace{-1mm}
\end{figure*}

\begin{algorithm}[tb]
  \caption{Extended MES}
  \label{alg:Extended MES}
  \begin{algorithmic}
    \INPUT data subset $\mat X \in \set{X}^N$, $n$, classifier $\decfunc$, input density $p$
    \REPEAT
    \STATE $\vec x \assign$ random point from $\mat X$
    \STATE $\data \assign$ $n$ samples from $p''$ (see~\eqref{eq:ppp def}) using $\decfunc$, $\vec x$, and $p$
    \STATE Set $\theta^*$ by fitting linear SVM (or logistic reg.) to $\data$
    \STATE Delete from $\mat X$ points correctly classified by SVM
    \STATE Append fitted parameters to list $L$
    \UNTIL $\mat X$ empty
    \OUTPUT parameter list $L$ (used for $\explr_{1:\nexplr}$)
  \end{algorithmic}
\end{algorithm}

Extended MES is based on upon a two-phase approach.
We first find the parameters for our explanations $\explr_{1:\nexplr}$ using Algo.~\ref{alg:Extended MES}.
Since the methods of Section~\ref{sec:Black box models} have finite sample guarantees, the output of Algo.~\ref{alg:Extended MES} is passed to Algos.~\ref{alg:MES MC Precomputation} and~\ref{alg:Run MES} to provide the final explanations.

\section{Face recognition example}
\label{sec:Face recognition example}

We now demonstrate MES on the scikit-learn demo ``Faces recognition example using eigenfaces and SVMs.''
The faces are reduced to dimension $D\!=\!150$ from $50 \times 37 \!=\! \textrm{1,850}$ using PCA\@.
Then 966 training examples are plugged into a (Gaussian kernel) multiclass SVM for classifying the faces as one of seven political figures.
When explaining a classification of face $k$ (e.g., Bush) we convert the SVM to a binary black box, informally as \smash{$\decfunc(\vec x) = \I\{\textrm{SVM}(\vec x) = k\}$}.
Throughout this paper, we use $\epsilon=0.025$ and $\delta=0.05$ implying $n=\textrm{129,099}$.
Induced from the assumptions of PCA, we use a standard multivariate Gaussian for the input density~$p(\vec x)$.

\citet{Turner2015} showed how to use standard MES to explain why the SVM classifies Hugo Chavez as George W Bush.
Here, we are also able to find interesting explanations using the linear explanations from Section~\ref{sec:extended MES}.
In Fig.~\ref{fig:linear face opt} we show a correct prediction of Colin Powell, and use MES to shed light on the responsible elements of the images.
Extended MES allows the explanation faces on the right in Fig.~\ref{fig:linear face opt} to be any image, not just an eigenface as was the case with standard MES and axis aligned explanations.

In Fig.~\ref{fig:linear face opt}, think of the white areas in the far right image as being the parts of the image that contribute to the SVM predicting Powell, and the dark areas as though the Powell prediction is made in spite of them.
Matches between the input face and explanation face of black $\times$ black or white $\times$ white positively contribute to the prediction of the classifier $\decfunc$, and white $\times$ black negatively contributes to the classification.
Patterns in the explanation face can be thought of as a sort of ``linear template.''
If the input face matches them exactly it leads to a large positive contribution.

Interpreting Fig.~\ref{fig:linear face opt}, we see that the SVM is ``picking up'' on the dark shading on the left side of Powell's chin, shading below his left eye, and a wide area for the dark pixels of his nostrils and nasolabial folds (smile lines)\@.
Indeed, in many training images of Powell the lighting is to his right.
MES has uncovered the high relevance that the classifier places on these non-obvious features.

\section{Credit scoring example}
\label{sec:Credit scoring example}
To further show the generality of MES, we use it on the UCI German credit data set.
After encoding the categorical data, there is a total of 48 possible features.
We chose to apply MES to $L_1$ logistic regression (LR) as it was the top performing model after an extensive comparison including SVMs and decision trees.
For the input distribution, we use the empirical distribution on the training data.
For simplicity we use axis aligned explanations with Algos.~\ref{alg:MES MC Precomputation} and~\ref{alg:Run MES}.

\begin{figure}
  \centering
  \includegraphics[scale=1.0]{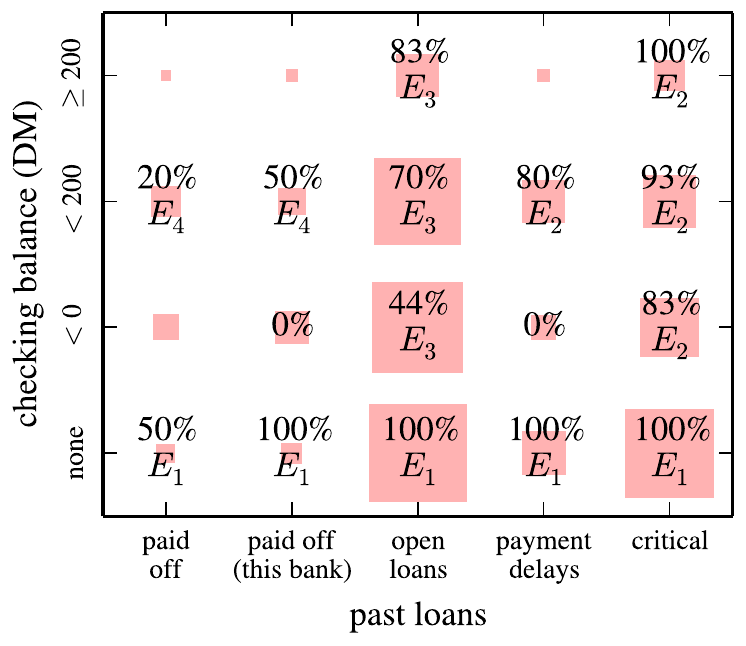}
  \caption{{\small
  MES applied to German credit data with LR classifier $\decfunc$.
  The shaded boxes represent the marginal distribution on the two variables (past loans and checking balance)\@.
  The area is proportional to the frequency in the training data.
  The percentages show how often test points with those values result in a classification of 1 by $\decfunc$.
  We show the \emph{most common} explanation for data points in each box.
  The explanations within a box vary as there are another 18 features not plotted.
  The explanations are:
  $E_1$ individual has no checking account;
  $E_2$ past payment delays or worse;
  $E_3$ individual already has loans out;
  $E_4$ loan duration less than 22 months.
  It is unclear why shorter loans are more likely to be predicted as risky by the model.
  However, $E_4$ is only used $1\%$ of the time and for individuals who are otherwise low risk.}}
  \label{fig:credit}
\end{figure}

The explanations for $99\%$ of the test set data points use either the feature ``credit history'' or ``status of existing checking account.''
The remaining $1\%$ of explanations use the loan duration feature.
Hence, in Fig.~\ref{fig:credit} we demonstrate the output of MES on data points in the cross section of credit history and checking account status.
The four explanations found in the test set have scores: 0.491 ($E_1$), 0.275 ($E_2$), 0.256 ($E_3$), and 0.244 ($E_4$).

The $L_1$ penalty also deems credit history and checking balance to be the most important features; only these two remain when the regularization penalty is increased.
However, constraining LR to only use these two features results in a model that disagrees with the predictively optimal model on $22.4\%$ of the test points.

\vspace{-2.5mm}
\section{Conclusions}
\vspace{-1mm}
We have presented a general framework for explaining black box models.
It alleviates the tension between performance and interpretability.
We described a new MC algorithm that finds explanations with many free parameters.

\bibliographystyle{icml2016}
\bibliography{MESreferencesMLSP}

\end{document}

%% file: mathmacros.tex
\usepackage{amsmath}
\usepackage{amsfonts}
\usepackage{amsthm}

\newcommand{\set}[1]{\mathcal{#1}}
\newcommand{\field}[1]{\mathbb{#1}}
\newcommand{\R}{\field{R}} 

\renewcommand{\vec}[1]{{\boldsymbol{\mathbf{#1}}}} 
\newcommand{\mat}[1]{{\boldsymbol{\mathbf{#1}}}} 

\newcommand{\E}{\field{E}}

\newcommand{\T}[0]{^{\top}} 
\newcommand{\TT}[0]{^{\!\raisebox{-0.6ex}{$\top$}}}

\newcommand{\hada}{\odot} 

\newcommand{\argmax}{\operatornamewithlimits{argmax}}
\newcommand{\argmin}{\operatornamewithlimits{argmin}}
\newcommand{\I}{\mathbb{I}}

\newcommand{\sample}{\sim}
\newcommand{\given}{|}
\newcommand{\data}{\mathcal{D}} 

\newcommand{\cov}[1]{\textrm{Cov}\left[#1\right]}

\DeclareMathOperator{\assign}{\leftarrow}